\pdfoutput=1

\documentclass[11pt]{article}

\usepackage[final]{acl}

\usepackage{times}
\usepackage{latexsym}

\usepackage[T1]{fontenc}

\usepackage[utf8]{inputenc}

\usepackage{microtype}

\usepackage{inconsolata}
\usepackage{amsfonts}
\usepackage{amsmath}
\usepackage{array}
\usepackage{natbib}
\usepackage{makecell}
\usepackage{graphicx}
\usepackage{subcaption}
\usepackage{hyperref}
\usepackage{multirow}
\usepackage{multicol}
\usepackage{booktabs}
\usepackage{enumitem}
\usepackage{stfloats}
\usepackage{pgfplots}
\usepackage{pgfplotstable}
\usepackage{tcolorbox}
\tcbuselibrary{skins}
\pgfplotsset{compat=newest}
\usepgfplotslibrary{colorbrewer}
\tcbset{
    enhanced, 
    colback=white, 
    colframe=black, 
    coltitle=white, 
    fonttitle=\bfseries, 
    colbacktitle=black, 
    toptitle=1mm, 
    bottomtitle=1mm, 
    left=3mm, 
    right=3mm, 
    top=3mm, 
    bottom=3mm, 
    sharp corners, 
    rounded corners=north, 
    boxrule=0.5mm, 
    arc=3mm 
}

\title{LLM with Relation Classifier for Document-Level Relation Extraction}
\author{Xingzuo Li\textsuperscript{\rm $1$}, Kehai Chen\textsuperscript{\rm $1$}, Yunfei Long\textsuperscript{\rm $2$} and Min Zhang\textsuperscript{\rm $1$} \\
$\textsuperscript{\rm $1$}$School of Computer Science and Technology, Harbin Institute of Technology, Shenzhen, China \\
$\textsuperscript{\rm $2$}$School of Computer Science and Electronic Engineering, University of Essex, Colchester, United Kingdom \\
\texttt{24S051028@stu.hit.edu.cn}, \texttt{chenkehai@hit.edu.cn}, \\
 \texttt{yl20051@essex.ac.uk}, \texttt{zhangmin2021@hit.edu.cn}
}

\begin{document}
\maketitle
\begin{abstract}
Large language models (LLMs) have created a new paradigm for natural language processing. Despite their advancement, LLM-based methods still lag behind traditional approaches in document-level relation extraction (DocRE), a critical task for understanding complex entity relations within long context. This paper investigates the causes of this performance gap, identifying the dispersion of attention by LLMs due to entity pairs without relations as a key factor. 
We then introduce a novel classifier-LLM approach to DocRE. Particularly, the proposed approach begins with a classifier designed to select entity pair candidates that exhibit potential relations and then feed them to LLM for final relation classification. This method ensures that the LLM's attention is directed at relation-expressing entity pairs instead of those without relations during inference. Experiments on DocRE benchmarks reveal that our method significantly outperforms recent LLM-based DocRE models and narrows the performance gap with state-of-the-art BERT-based models.\footnote{Our code is publicly available at: \url{https://github.com/wisper12933/LMRC}.}

\end{abstract}

\section{Introduction}

Document-level relation extraction (DocRE) is a critical yet challenging task within the broad field of natural language processing (NLP). It involves identifying and extracting semantic relationships between entity pairs across an entire document, rather than confining the analysis to individual sentences. This task poses significant challenges due to the coreference resolution and long-distance dependencies inherent in the textual corpus. Traditional DocRE models have relied on sophisticated neural network architectures to simulate human-like comprehending and reasoning processes, such as GNNs \cite{graph_1, graph_2} and attention networks \cite{attention_3, attention_2}, achieving state-of-the-art performance \cite{method_4}.
Recently, the robust semantic comprehension and generation capabilities of large language models (LLMs) have emerged as a promising avenue for advancing DocRE. Due to extensive pre-training, LLMs possess a broader coverage of knowledge across various fields and demonstrate enhanced versatility. With minimal fine-tuning, they can quickly adapt to domain-specific tasks. Besides, the textual output and in-context learning abilities of LLMs enable them to accomplish tasks in a more interpretable manner. Building on these advantages, researchers \cite{autore, chatbased, DocGNRE} have intensively explored innovative strategies to further enhance the capabilities of LLMs in addressing complex DocRE tasks.

\begin{figure}
    \centering
    \includegraphics[width=2.5in, height=2.2in]{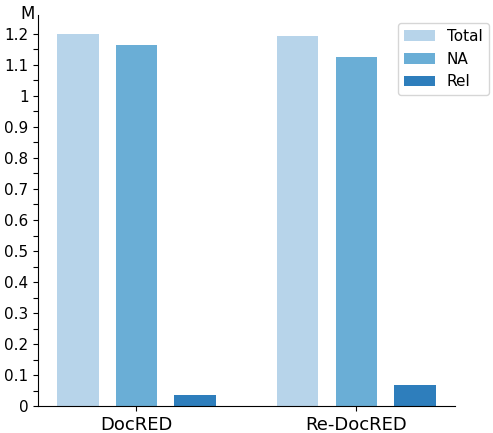}
    \caption{Statistics on the number of entity pairs in DocRED and Re-DocRED training set. \textit{NA}: entity pairs that do not express any relation. Rel: entity pairs expressing relations.}
    \label{fig:introduction_fig}
\end{figure}

Despite these efforts, the performance of LLM-based methods still fall short of traditional approaches. Our detailed analysis reveals a key issue: DocRE tasks necessitate the construction of an exhaustive set of potential entity pairs, which we refer to as \textbf{candidate space}. Within this large set of entity pairs, only a small subset expresses relations, leading to a \textbf{significant imbalance} in the candidate space. As shown in Figure \ref{fig:introduction_fig}, our preliminary investigations, conducted on two widely adopted benchmark datasets, DocRED \cite{method_2} and Re-DocRED \cite{Re-DocRED}, underscores this phenomenon. The overwhelming majority of entity pairs labeled as \textit{NA} (no relation), dominates the candidate space, potentially distracting LLMs from the minority that do encode meaningful relations. Consequently, this data distribution skew is recognized as a major factor contributing to the performance shortcomings exhibited by LLMs on DocRE.

Based on this finding, this paper introduces a novel method named LMRC (\textbf{LLM} with \textbf{R}elation \textbf{C}lassifier), which aims at bridging the performance gap between LLM-based DocRE methods and their conventional counterparts. Specifically, LMRC conceptualizes DocRE as a workflow comprising two stages: \textbf{Relation Candidate Proposal} (RCP) and \textbf{Relation Classification} (RC). In the RCP stage, a pre-processing classifier is applied, which explicitly harnesses attention mechanism to filter out \textit{NA} entity pairs that do not harbour any relation. This strategic filtering not only drastically reduces the size of candidate space but also ensures that the subsequent stage operates on a more relevant set of entity pairs. Subsequently, the RC stage leverages the remarkable capabilities of LLMs to perform multi-classification within this refined candidate space, ensuring a more focused and accurate relation extraction process. Experimental results on the DocRED and Re-DocRED benchmarks shows that our proposed LMRC methodology gains significant improvement over other LLM-based DocRE methods, suggesting its viability as a strategy for future DocRE.

Our main work and contributions are as follows:

\begin{itemize}[leftmargin=*, labelsep=0.7em, topsep=1pt, itemsep=2.5pt, parsep=1pt]
\item This study empirically found that the key reason why LLM-based methods struggle to perform well on DocRE tasks lies in the imbalance of the candidate space.

\item The proposed LMRC approach introduces a pre-classifier into LLMs to greatly narrow the performance gap with state-of-the-art BERT-based DocRE models. 

\item Empirical experiments conducted on DocRED and Re-DocRED datasets showed the feasibility and effectiveness of The proposed LMRC approach.
\end{itemize}

\section{Related Works}
\noindent\textbf{Document-level Relation Extraction} Traditional NLP methods for identifying relations in documents, such as CNNs and LSTMs \cite{method_2, sequence_1, sequence_2}, are built on the foundations of early deep neural networks. Specifically, Jia et al. \cite{method_6} advocated for an entity-centric, multiscale approach, enabling a singular relation prediction for each entity tuple. More recently, approaches for document-level relation extraction have evolved around two main perspectives: Graph-based and Transformer-based. The concept of document graph was initially introduced by Quirk and Poon \cite{graph_basic}. In this paradigm, words are represented as nodes, while linguistic relations (e.g., adjacency, syntactic dependencies) are captured as edges. 
Sun et al. \cite{graph_4} proposed a method that combines dual-channel and hierarchical graph convolutional network, utilizing GCN to gather pertinent information distributed across the document to enhance the inference of implicit relations. 
Peng et al. \cite{sbg} introduced a subgraph reasoning framework that simplifies various paths into a streamlined subgraph structure, facilitating various reasoning processes.
Specially, graph-based methods GAIN \cite{gain} and LSR \cite{lsr} excels in handling long-distant dependencies that often proved crucial in comprehending complex relations. 
In addition to modifying the graph structure, Xu et al. \cite{sief} proposed a sentence importance estimation and focusing framework, which encourages models to focus on evidence sentences.
Among early Transformer-based methods, Verga et al. \cite{trans_1} introduced Bi-affine Relation Attention Network to identify the relations between entities. 
Han and Wang \cite{trans_2} utilized a document-level entity mask method augmented with type information to provide a more richer contextual understanding of the entities.
Yu et al. \cite{rsman} accounted for the importance of interactions among mentions and used a relation-specific mention attention network to incorporate relation into prototype representations.
Zhou et al. \cite{method_1} and Tan et al. \cite{attention_2} further leveraged attention mechanism to enhance entity representations, ultimately improving the accuracy of relation classification.
Recently, Ma et al. \cite{method_4} extended previous research by integrating knowledge distillation strategy and achieved SOTA performance.

\noindent\textbf{Relation Extraction with LLMs} \hspace{0.2em} Several recently proposed RE approaches have advocated the utilization of conditional generative models to output string encodings, thereby addressing the task in an interpretable manner. Paolini et al. \cite{add_1} introduced a framework that reframes many structured prediction tasks, including relation extraction, as a sequence-to-sequence problem. Wan et al. \cite{add_5} pushed the boundaries by prompting GPT-3 for relation extraction on datasets such as TACRED \cite{introduction_4} and SciERC \cite{scierc}. Wadhwa et al. \cite{introduction_7} extended this line of work by leveraging few-shot in-context learning for sentence-level relation extraction. They also evaluated Flan-T5 to prove the potential of fine-tuning smaller scale models for accomplishing this challenging task.

Previously, the application of LLMs to document-level relation extraction has been limited. 
Ozyurt et al. \cite{REPLM} introduced a method for in-context few-shot relation extraction leveraging pre-trained language models, providing a viable paradigm for tackling DocRE with LLMs. 
Xue et al. \cite{autore} guided LLMs to extract triplets from documents using a pipeline approach, starting with the extraction of head entity, followed by identifying the relation, and concluding with the retrieval of tail entity.
Sun et al. \cite{chatbased} attempted to employ chain-of-retrieval and denoising strategy to steer LLMs in understanding relations and generating high-quality synthetic data. During their research on GPT-3's contextual learning abilities for document-level biomedical information extraction tasks, Gutierrez et al. \cite{add_4} uncovered a general issue in in-context learning when dealing with the \textit{NA} category. Inspired by this discovery, we first focus on selecting potential relation-expressing entity pairs and then utilize the reasoning capabilities of LLMs to complete relation classification. As far as we know, our work is among the first to apply LLMs for document-level relation extraction tasks.

\noindent\textbf{In-Context Learning} \hspace{0.2em} In-context learning \cite{ICL} has emerged as a competitor against supervised baselines across a wide
range of tasks including semantic parsing, machine translation and question answering \cite{prompt_retrieval, icl_add1, icl_add2}. This approach circumvents the computational burden of weight updates by leveraging a handful of pertinent examples presented to the model. Recent works on in-context learning have mainly focused on optimizing prompt design \cite{p_design}, prompt ordering \cite{prompt_order} and prompt retrieval \cite{prompt_retrieval, retrieval_2}. Typically, Wan et al. \cite{add_5} and Wadhwa et al. \cite{introduction_7} have conducted comprehensive evaluations of the in-context learning capabilities of GPT-3 on sentence-level RE tasks. We conduct our experiment in \ref{sec:gpt} following their pioneering works.

\section{Preliminary}\label{subsec:2}

\subsection{Problem Fomulation}
Given a document $D$ that includes a set of sentences $X_D={\{x_i\}}_i^k$ and a set of entities $E_D={\{e_i\}}_i^n$, document-level relation extraction aims to predict a subset of relations from $R \cup \{NA\}$ for the entity pairs ${(e_s, e_o)}_{s,o=1,...,n;s\neq o}$, where $R$ represents a predefined set of relations, $e_s$ and $e_o$ respectively denote the subject entity and the object entity, and \textit{NA} indicates that there is no relation between the entities. A entity $e_i$ can appear multiple times within a document through its mentions $M_{i}={\{m_j^i\}}^{N_{i}}_{j=1}$, where $m_j^i$ represents the $j^{th}$ mention and $N_{i}$ denotes the number of mentions. During test time, the model is required to predict labels for all entity pairs.

\subsection{Pre-Experiments}
To investigate the reasons behind the under-performance of current LLMs in DocRE, we directly fine-tune LLaMA2-13B-Chat and report the outcomes and key insights derived from applying this approach to DocRED and Re-DocRED datasets.

\noindent\textbf{Fine-tuning LLaMA2}
In order to construct effective prompts for this task, we employ the following instructional guideline: \textit{Your task is to determine whether there are relations between the entity pairs based on the information in the text. If there exist relations, select relations for the entity pairs from the relation set; if there is no relation, return None}. The instruction is accompanied by an input that includes a predefined relation set, the text corresponding to the document, and the entity pairs that need to be classified. To prevent ambiguities and optimize token usage in our process, we use \textit{None} as a substitute for \textit{NA} and require the model to explicitly label entity pairs that do not contain relations as \textit{None}. The complete prompt format is provided in Appendix \ref{box:prompt_re}. 

Each document in DocRE task involves a considerable number of tokens, frequently exceeding the maximum limit supported by typical processing unit. To tackle this challenge, we devise a method to conduct relation extraction for each document $D$ via $\frac{n \times (n - 1)}{k}$ inputs, where $n$ denotes the number of entities within document $D$, and the variable $k$ represents the capacity limit for entity pairs per input instance. We then integrate all entity pairs into the inputs adhering to the prescribed scheme to perform LoRA \cite{lora} fine-tuning and testing. In addition, entities of the triplets returned by models are aligned to the annotations in the dataset using \href{https://github.com/seatgeek/thefuzz}{thefuzz}, and the relations generated not in the predefined relation set are considered incorrect.

\noindent\textbf{Results} \hspace{0.2em}
Statistics of DocRED and Re-DocRED are shown in Table \ref{tab:statistics}. \textit{NA} entity pairs, which do not exhibit relations, constitute a significant proportion in both datasets, leading to an imbalance in the candidate space. Further to the empirical observations by \cite{autore}, our study extends the analysis to examine the model's outputs from a distribution perspective, supported by experiments, aiming to uncover the fundamental reasons behind the observed under-performance. As demonstrated in Table \ref{tab:pre}, the number of triples generated by LLaMA2-13B-Chat is markedly less than that annotated in the dataset. This discrepancy suggests that LLMs tend to label relation-expressing entity pairs as \textit{NA}, leading to poor outcomes.

\begin{table}[t]
\centering
{\scriptsize
\begin{tabular}{lcccc}
\toprule
\multirow{2}{*}{\textbf{Description}} & \multicolumn{2}{c}{\textbf{DocRED}} & \multicolumn{2}{c}{\textbf{Re-DocRED}}\\
&\textbf{Dev}&\textbf{Test}&\textbf{Dev}&\textbf{Test}\\
\midrule
Candidate Space            &   395,572     & 392,158 & 193,232 & 198,670 \\
\# \textit{NA} Entity Pairs       &   384,949     & -       & 179,870 & 185,043 \\
\# Relation Entity Pairs   &   10,623      & -       & 13,362  & 13,627  \\
\# Annotated Triples       &   12,275      & -       & 17,284  & 17,448  \\
\bottomrule
\end{tabular}
}
\caption{Statistics on DocRED and Re-DocRED}
\label{tab:statistics}
\end{table}

\begin{table}[t]
\centering
{\small
\begin{tabular}{lcccc}
\toprule
\multirow{2}{*}{\textbf{Metrics}} & \multicolumn{2}{c}{\textbf{DocRED}} & \multicolumn{2}{c}{\textbf{Re-DocRED}} \\
& \textbf{Dev} & \textbf{Test} & \textbf{Dev} & \textbf{Test} \\
\midrule
$Precision$         &  69.00       & -       &  84.88  &  83.94\\
$Recall$            &  27.43       & -       &  38.06  &  38.14\\
$F_1$               &  39.25       &  38.66  &  52.56  &  52.45\\
Ign $F_1$           &  38.62       &  38.09  &  52.29  &  52.15\\
\# Extracted Triples&  4,925       &  4,932  &  7,787  &  7,979\\
\bottomrule
\end{tabular}
}
\caption{Results of preliminary experiment.}
\label{tab:pre}
\end{table}

\section{Proposed Method: LMRC}
\begin{figure*}[pt]
    \centering
    \includegraphics[width=0.9\linewidth]{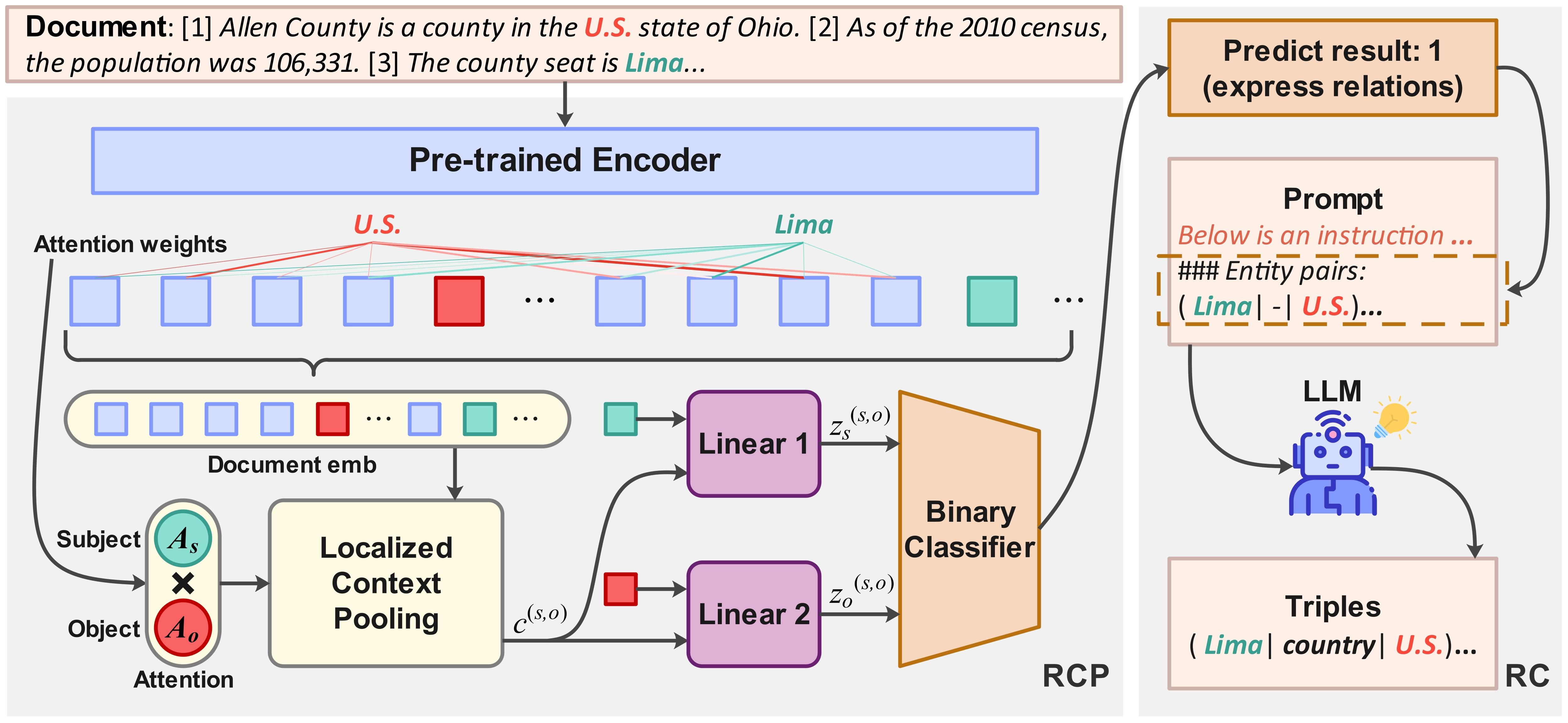}
    \caption{\textbf{Illustration of LMRC}. Relation Candidate Proposal(RCP) leverages localized context pooling \cite{method_1} in the construction of a pre-processing classifier, focusing on selecting entity pairs that contain relations. Relation Classification(RC) takes the results from the previous stage to create prompt that guides fine-tuned LLaMA2 to accomplish multi-classification task.}
    \label{fig:long}
\end{figure*}

To prevent LLMs from prioritizing \textit{NA} entity pairs, the proposed LMRC methodology, as shown in Figure \ref{fig:long}, initially uses traditional neural networks for \textbf{Relation Candidate Proposal} to identify relation-expressing entity pairs. Then, LLMs rely on these proposals for \textbf{Relation Classification}.

\subsection{Relation Candidate Proposal}
In this stage, we build a simple model to conduct a binary classification task, with the outcome being entity pairs expressing relations. As prior works \cite{attention_2, method_4} have shown that contextual information is indispensable for the relation extraction task, our model adapts localized context pooling from \citet{method_1}.

\vspace{7pt}

\noindent \textbf{Entity Representation} \hspace{0.3em} Following the entity marker technique \cite{introduction_4, method_5}, a special token "*" is inserted at the start and end position of each entity mention. Then, tokens $T=\{t_i\}_{i=1}^l$ within document $D$ are encoded by a Transformer-based\cite{transformer} pretrained language model (PLM) to generate contextualized embeddings $\mathbf{H}$ along with their attentions $\mathbf{A}$:
\begin{equation}\label{eq:eq_1}
    \mathbf{H},\mathbf{A}=PLM(T), 
\end{equation}
where $\mathbf{H} \in \mathbb{R}^{l \times d}$, $\mathbf{A} \in \mathbb{R}^{H \times l \times l}$, $d$ is the hidden dimension of the PLM and $H$ is the number of attention heads. We take the embedding of "*" at the start of mentions as their embeddings. The entity embedding $h_{i} \in \mathbb{R}^d$ for each entity $e_i$ with mentions $M_{i}=\{m_j^i\}_{j=1}^{N_{i}}$ is computed by logsumexp pooling \cite{method_6}:
\begin{equation}
    h_{i}=log\sum_{j=1}^{N_{i}}exp(h_{m_j^i}).
\end{equation}

\vspace{2pt}

\noindent \textbf{Localized Context Representation} \hspace{0.3em} For each entity $e_i$, we aggregate the attention output for its mentions by mean pooling $A_{i}=\sum_{j=1}^{N_{i}}(a_{m_j^i})$, where $a_{m_j^i}\in \mathbb{R}^{H\times l}$ is the attention weight at the position of mention $m_j^i$ from the last layer. Then given an entity pair $(e_s,e_o)$, its localized context embedding $c^{(s,o)}\in \mathbb{R}^d$ can be obtained by:
\begin{align}
    q^{(s,o)}&=\sum_{i=1}^H (A_{s}^i\circ A_{o}^i), \\
    c^{(s,o)}&= \mathbf{H}^\top q^{(s,o)},
\end{align}
where $q^{(s,o)}\in \mathbb{R}^l$ is the mean-pooled attention weight for entity pair $(e_s,e_o)$ and $\mathbf{H}$ is the contextualized embedding in Eq.(\ref{eq:eq_1}).

\vspace{7pt}

\noindent \textbf{Binary Classification} \hspace{0.3em} To predict whether entity pair $(e_s,e_o)$ expresses relations, we first generate context-enhanced entity representations: 
\begin{align}
    z_s^{(s,o)}=tanh(\mathbf{W}_s h_{s}+\mathbf{W}_c c^{(s,o)}),
\end{align}
where $\mathbf{W}_s,\mathbf{W}_c\in \mathbb{R}^{d\times d}$ are trainable parameters. We obtain the object representation $z_o^{(s,o)}$ in the same manner. Then, a bilinear classifier is applied on the representations to compute the probability:
\begin{equation}
    P(\textit{NA}|e_s,e_o)=\sigma(z_s^{(s,o)\top}\mathbf{W} z_o^{(s,o)} + b),
\end{equation}
where $\mathbf{W} \in \mathbb{R}^{d\times d}$ is a trainable parameter matrix, $\sigma$ is the sigmoid function, $P(\textit{NA}|e_s,e_o)$ is the probability that entity pair $(e_s,e_o)$ does not express any relation. We choose Binary Cross Entropy as our loss function:
{\small
\begin{equation}
\mathcal{L}=\sum I(\textit{NA})logP+(1-I(\textit{NA}))log(1-P),
\end{equation}}
where $I$ represents Indicator Function, and $I(\textit{NA})$ takes the value of 1 when there is no relation between entities. $P$ denotes $P(\textit{NA}|e_s, e_o)$.
\subsection{Relation Classification}
\label{sec:rc}

After identifying potential relation-expressing entity pairs in the RCP stage, we employ the method described in Section \ref{subsec:2}, leveraging LLMs to undertake supervised fine-tuning for relation classification. This approach involves a subtle modification to the previous prompting scheme, where we streamline the process by eliminating the \textit{NA} category and change certain expressions. These adjustments are intended to further sharpen model's concentration on the classification task, ensuring a more precise and targeted performance. Details can be found in Appendix \ref{box:prompt_rc}.

Additionally, the optimization achieved through the RCP stage yields a substantial reduction in the number of inputs per document, transitioning from a complexity of $\frac{n \times (n - 1)}{k}$ to $\frac{n'}{k}$, where $n'$ denotes the number of entity pairs identified in the RCP stage. This reduction, driven by the elimination of \textit{NA} instances, not only enhances computational efficiency but also shortens the inference time. 

\section{Experiments}

\subsection{Experimental Settings}
\noindent \textbf{Datasets} \hspace{0.3em} We conduct experiments on DocRED \cite{method_2} and Re-DocRED \cite{Re-DocRED}, two large-scale crowd-sourced benchmark datasets tailored for document-level RE. 

DocRED, rooted in Wikipedia and Wikidata, stands as a milestone in DocRE. It provides a comprehensive annotation schema, encompassing entity mentions and types, relational facts, as well as the corresponding supporting evidence. With 97 distinct target relations and an average of approximately 26 entities per document, DocRED offers a rich and challenging landscape for researchers. Notably, over 40.7\% of the relational facts within DocRED require multi-sentence reasoning. Although DocRED is a widely recognized benchmark, the annotations of the dataset remain incomplete. \citet{Re-DocRED} thus proposed Re-DocRED, a more reliable benchmark for DocRE that revises DocRED to mitigate the false negative issue within it.

\vspace{5pt}

\noindent \textbf{Configuration} \hspace{0.3em} In the RCP stage, RoBERTa$_{\mathrm{large}}$ \cite{roberta} is chosen as the foundational PLM encoder. We implement early stopping based on the $F_1$ score obtained from the development set. We adopt AdamW \cite{adamW} as the optimizer and apply a linear warmup for the learning rate at the first 6\% steps. Main parameters are provided in Table \ref{tab:hyper_RCP} in Appendix \ref{sec:para}. We use development set to manually tune the optimal parameters for the RCP stage based on the F1 score. The value of parameters we finally adopted are in bold. 

In the RC stage, we fine-tune LLMs with the RC-specific prompt, which is detailed in \ref{sec:rc}, using LoRA. Details regarding parameters are provided in Table \ref{tab:hyper_lora} in Appendix \ref{sec:para}. We train and evaluate the pre-classifier on a single NVIDIA A100 GPU during the RCP stage, and fine-tune LLMs on 4 NVIDIA A100 GPUs during the RC stage.

\vspace{5pt}

\noindent\textbf{Evaluation} \hspace{0.2em} Consistent with other SOTA methods, we apply the standard evaluation metrics: $F_1$ and Ign $F_1$. Ign $F_1$ is calculated by excluding triplets that are already present in the training set from both the development and test sets.

\subsection{Main Results}

\begin{table*}[ht]
    \centering
    \begin{subtable}[t]{0.6\textwidth}
        \centering
        \scalebox{.77}{
        \footnotesize
        \begin{tabular}{lcccc}
            \toprule
            \multirow{2}{*}{\textbf{Method}} & \multicolumn{2}{c}{\textbf{Dev}} & \multicolumn{2}{c}{\textbf{Test}} \\
                     & Ign $F_1$  & $F_1$ & Ign $F_1$ & $F_1$ \\
            \midrule
            \textbf{BERT-based} \\
             HIN-BERT$_{base}$\cite{hinbert} & 54.29 & 56.31 & 53.70 & 55.60 \\
             CorefBERT$_{base}$\cite{coref} & 55.32 & 57.51 & 54.54 & 56.96 \\
             CorefRoBERTa$_{large}$\cite{coref}& 57.35 & 59.43 & 57.90 & 60.25 \\
             HeterGSAN+reconstruction\cite{Heter_re} & 58.13 & 60.18 & 57.12 & 59.45 \\
             SSAN-RoBERTa$_{large}$\cite{ssan} & 60.25 & 62.08 & 59.47 & 61.42 \\
             KD-RoBERTa$_{large}$\cite{attention_2}& 65.27 & 67.12 & 65.24 & 67.28 \\
             DREEAM-RoBERTa$_{large}$\cite{method_4}& 65.52 & 67.41 & 65.47 & 67.53 \\
            \midrule
            \textbf{LLM-based} \\
             0-shot GPT-4\cite{result_add}& - & -          & - & 15.58 \\
             MDP GPT-4\cite{result_add}   & - & -          & - & 21.33 \\
             FT Flan-UL2\cite{result_flan}& - & -          & - & 54.50 \\
             REPLM GPT3.5\cite{REPLM}     & - & 59.66      & - & -    \\
            \midrule
            \textbf{Our Methods} \\
            LoRA FT LLaMA2-7B-Chat$^{\dag}$   & 33.95 & 34.32 & 33.99 & 34.34 \\
            LoRA FT LLaMA2-13B-Chat$^{\dag}$  & 38.62 & 39.25 & 38.09 & 38.66 \\
            LMRC LLaMA2-7B-Chat               & 52.40 & 54.10 & 52.81 & 54.73 \\
            LMRC LLaMA2-13B-Chat              & 58.16 & 59.97 & 58.49 & 60.52 \\
            LMRC LLaMA3-8B-Instruct           & 59.13 & 61.01 & 58.77 & 60.84 \\
            LMRC LLaMA3.1-8B-Instruct         & 59.22 & 61.08 & 58.62 & 60.69 \\
            \bottomrule
        \end{tabular}
        }
        \caption{Results on the development and test set of DocRED.}
        \label{tab:docred}
    \end{subtable}%
    \begin{subtable}[t]{0.4\textwidth}
        \centering
        \scalebox{.77}{
        \footnotesize
        \begin{tabular}{lcc}
            \toprule
            \textbf{Method} & Ign $F_1$ & $F_1$ \\
            \midrule
            \textbf{BERT-based} \\
            KD-RoBERTa$_{large}$\cite{attention_2} & 77.60  & 78.28 \\
            DREEAM\cite{method_4} & 79.66  & 80.73 \\
            \midrule
            \textbf{LLM-based} \\
            0-shot GPT-4\cite{result_add}     & - & 14.21 \\
            MDP GPT-4\cite{result_add}        & - & 20.16 \\
            AutoRE-Vicuna-7B\cite{autore}     & - & 53.84      \\
            \midrule
            \textbf{Our Methods} \\
            LoRA FT LLaMA2-7B-Chat$^{\dag}$  &  52.74 &  53.02 \\
            LoRA FT LLaMA2-13B-Chat$^{\dag}$ &  52.15 &  52.45 \\
            LMRC LLaMA2-7B-Chat              &  72.33 &  72.92 \\
            LMRC LLaMA2-13B-Chat             &  74.08 &  74.63 \\
            LMRC LLaMA3-8B-Instruct          &  72.21 &  72.80 \\
            LMRC LLaMA3.1-8B-Instruct        &  72.35 &  72.93 \\
            \bottomrule
        \end{tabular}
        }
        \caption{Results on the test set of Re-DocRED}
        \label{tab:redocred}
    \end{subtable}
    \caption{Evaluation results on the DocRED and Re-DocRED datasets. The scores of prior methods are borrowed from corresponding papers. Results marked with $\dag$ are our baselines.}
    \label{tab:main_result}
\end{table*}

We compare our LMRC with pretrained BERT-based and LLM-based methods on both datasets. BERT-based methods, known for achieving state-of-the-art (SOTA) performance, utilize BERT family pretrained models as encoders. Recently introduced LLM-based methods employ fine-tuning, in-context learning with CoT \cite{introduction_8}, or retrieval augmented generation (RAG, \citet{rag}) to enhance the performance of LLMs on relation extraction.

\vspace{5pt}

\textbf{Overall Performance} \hspace{0.2em} As demonstrated in Table \ref{tab:docred}, the direct fine-tuning approach applied to LLaMA models and other LLM-based methods expose suboptimal performance and inefficiencies in processing capabilities, underscoring the inherent challenges faced when leveraging LLMs for DocRE. Our results also corroborate the findings of \cite{autore}. However, following the implementation of our proposed task division strategy, we observe a substantial improvement, marked by a significant boost in the $F_1$ score. Table \ref{tab:redocred} presents a comparative analysis of LMRC's performance against existing methods evaluated on the Re-DocRED test set. We observe that LMRC surpasses other LLM-based methods. Moreover, compared to the poor performance of recent LLM-based methods, our LMRC framework achieves higher scores across a broader range of evaluations, narrowing the performance gap with the current SOTA method, DREEAM, and positioning it as a promising paradigm for future DocRE. 

\begin{table}[t]
\centering
{\footnotesize
\begin{tabular}{lcc}
\toprule
\textbf{Method} & \textbf{Intra} & \textbf{Inter} \\
\midrule
BERT-RE$_{base}^{\dag}$    & 61.61 & 47.15 \\
RoBERTa-RE$_{base}^{*}$    & 65.65 & 50.09 \\
LSR-BERT$_{base}^{\dag}$   & 65.26 & 52.05 \\
GAIN-BERT$_{base}^{*}$     & 67.10 & 53.90 \\
\midrule
LoRA FT LLaMA2-7B-Chat$^{\dag}$  &  52.74 &  53.02 \\
LoRA FT LLaMA2-13B-Chat$^{\dag}$ &  52.15 &  52.45 \\
LMRC LLaMA2-7B-Chat              &  72.33 &  72.92 \\
LMRC LLaMA2-13B-Chat             &  74.08 &  74.63 \\
LMRC LLaMA3-8B-Instruct          &  72.21 &  72.80 \\
LMRC LLaMA3.1-8B-Instruct        &  72.35 &  72.93 \\
\bottomrule
\end{tabular}
}
\caption{Intra- and Inter-$F_1$ on the development set of DocRED. $\dag$ denotes results from \citet{lsr}, and * denotes results from \citet{gain}.}
\label{tab:table_intra_inter}
\end{table}

\vspace{5pt}

\textbf{Performance of Intra-/Inter-Extraction} \hspace{0.2em} We also report Intra-$F_1$/Inter-$F_1$, which respectively evaluate the model's capability in capturing intra- and inter-sentence relations. LSR \cite{lsr} and GAIN \cite{gain}, both graph-centric methodologies that leverages graph neural networks for inference, serve as valuable benchmarks. As illustrated in Table \ref{tab:table_intra_inter}, LMRC not only outperforms our baseline obtained by directly fine-tuning LLaMA models in both Intra- and Inter-$F_1$, but also remains competitive with advanced graph-based models like GAIN-BERT$_{base}$. This demonstrates LMRC's effectiveness, particularly in scenarios that require complex relational parsing within document-level context.

\subsection{Ablation Studies}
After demonstrating the effectiveness of our proposed LMRC, we further explore the roles of RCP and RC stages on the DocRED development set. To validate the RCP stage, we fine-tune LLaMA2-13B-Chat to replace the original pre-classifier for binary classification. For the RC stage, we introduce a more simplified setup where we feed the model with entity pairs annotated in the ground truth. By masking relation tags, we challenge the model to classify these pairs into the predefined relation set, fine-tuning LLaMA2-13B-Chat specifically for relation classification.

As shown in Table \ref{tab:rc_stage}, substituting our pre-classifier with the fine-tuned LLaMA2-13B-Chat significantly reduced the $F_1$ score of the RCP stage as well as the overall extraction performance. We argue that the intricate multi-relation nature and the discrete distribution of relations in DocRE tasks present unique challenges for LLMs, highlighting the crucial role of our pre-classifier in ensuring robust performance. Furthermore, the ablation study conducted on the RC stage highlights the effectiveness of fine-tuned LLaMA2-13B-Chat in performing relation classification. This phenomenon demonstrates the potential of LLMs to excel in complex relation extraction tasks.

\begin{table}[t]
\centering
{\footnotesize
\begin{tabular}{lccc}
\toprule
\textbf{Settings} & $F_1$ of RCP & Ign $F_1$ & $F_1$ \\
\midrule
\textbf{RCP stage}\\
LMRC                                 & 64.64 & 58.16 & 59.97 \\
\textit{w/o} pre-classifier \textit{w} LLM    & 31.30 & 23.22 & 24.59 \\
\midrule
\textbf{RC stage}\\
relation classification              &   -   & 86.09 & 86.75 \\
\bottomrule
\end{tabular}
}
\caption{Ablation studies evaluated on DocRED dev set.}
\label{tab:rc_stage}
\end{table}

\begin{figure*}[ht]
    \centering
    \includegraphics[width=\linewidth]{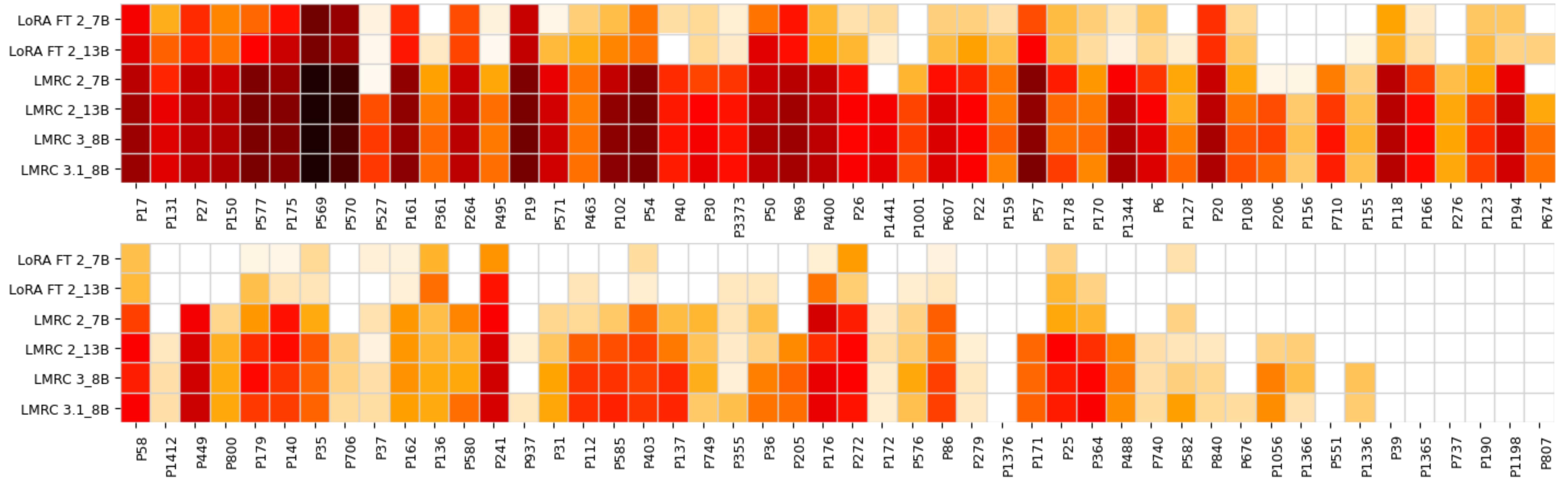}
    \caption{$F_1$ scores per relation type in the DocRED development set results (darker = better). White color means that no correct predictions were made for this relation. The relations are arranged in descending order by the number of triples.}
    \label{fig:heatmap_fig}
\end{figure*}


\subsection{Fine-grained Analysis for Each Relation}
\label{sec:fine-grain}
To further assess the model's sensitivity to different categories within the dataset, we undertakes a fine-grained statistical analysis of the $F_1$ score for each relation type in the DocRED development set, comparing both baseline and LMRC. Figure \ref{fig:heatmap_fig} visually showcases the enhancement effects in a more intuitive manner. 

Our meticulous data analysis leads to the following conclusions: (1) LMRC improves the extraction performance for the majority of relations. (2) A discernible positive relationship emerges between the frequency of relation occurrence in the training set and their corresponding extraction performance. High-frequency relations, such as P569: \textit{date of birth}, achieved an improved $F_1$ score exceeding 90, showcasing the model's proficiency in handling prevalent patterns. 

However, it is noteworthy that despite these positive results, LMRC encounters challenges in extracting instances pertained to certain relations (e.g., P39: \textit{position held}, P737: \textit{influenced by}). We consider that this limitation stems primarily from the scarcity of these relations in the training data, which restricts the model's capacity to sufficiently learn relevant features and nuances. Consequently, the model struggles to generalize effectively for these less frequently encountered relation types.

\subsection{Out-of-Domain Relation Studies}
In the aforementioned evaluation, we conservatively categorize all relations generated by LLMs that do not align with the predefined relation set as erroneous outcomes. However, previous research \cite{introduction_7} has indicated that evaluating LLM-based models should not entirely rely on exact matches to targets. This is because LLMs can generate outputs that are semantically proximate to the target relations but may not satisfy the stringent exact-matching requirement. For instance, while the phrase "\textit{works at}" in the model's output resembles "\textit{work for}" in the target, strict evaluation criteria would count it as a misclassification.

\begin{figure}[ht]
    \centering
    {
        \includegraphics[width=\linewidth]{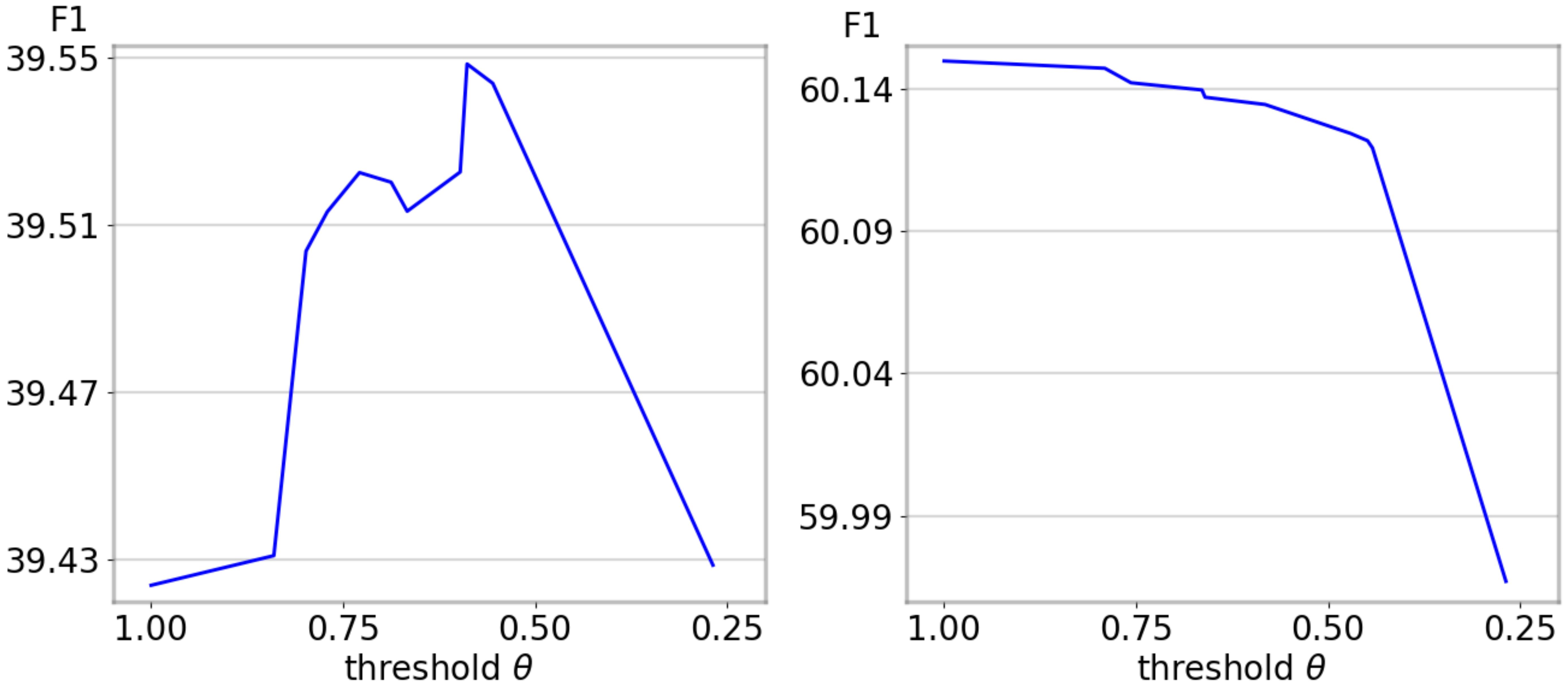}
    }
    \caption{The impact of threshold $\theta$ for cosine similarity on the $F_1$ score. Left: LoRA fine-tune LLaMA2-13B-Chat. Right: LMRC with LLaMA2-13B-Chat. Both methods are conducted on the DocRED dev set.}
    \label{fig:threshold}
\end{figure}

To gain a comprehensive understanding of this phenomenon, we delve deeper into the out-of-domain relations generated by LLaMA2-13B-Chat. We leverage SBERT \cite{sbert}, a well recognized approach for semantic similarity, to align out-of-domain relations to the predefined relation set $R$. This process first computes the cosine similarity between each out-of-domain relation and the relations within $R$. For each $r_i$ in the out-of-domain set, we choose the relation in $R$ that exhibits the highest similarity score $s_{\text{max}}^i$ as the optimal match. Intuitively, some out-of-domain relations may not carry meaningful information due to the hallucination issue of LLMs, and blindly aligning them to $R$ is not appropriate. Therefore, we introduce a heuristic threshold $\theta$. Only when $s_{\text{max}}^i\geq\theta$, does the alignment processed; otherwise, the triplets containing $r_i$ are discarded. After alignment, we recalculate the $F_1$ score of our methods on the DocRED development set. Results are provided in Figure \ref{fig:threshold}.

As is shown, LLaMA2-13B-Chat directly fine-tuned on DocRED attains its peak $F_1$ score when threshold $\theta$ is set approximately at 0.55. However, despite incorporating semantic alignment, the out-of-domain relations generated by LLaMA2-13B-Chat fine-tuned within the LMRC framework remain incorrect. We analyze these relations in depth and conclude that they mainly suffer from the following problems:
\begin{itemize}[leftmargin=*, labelsep=0.6em, topsep=1pt, parsep=1pt, itemsep=1pt]
\item The model outputs "-", which appears in the entity pair input format, as a relation.

\item While some of the generated out-of-domain relations could ostensibly be mapped to similar relations within set $R$, the classification result is incorrect.
\end{itemize}

Additionally, it is noteworthy that both methods generated a relatively low number of out-of-domain relations, 74 and 72, respectively. This limited occurrence helps explain why adjustments to the threshold $\theta$ has little impact on the $F_1$ score.

\subsection{Performance of GPT models}
\label{sec:gpt}
To ensure a thorough assessment, we expand our investigation to encompass a broader array of models, notably incorporating GPT-3-turbo-16k and GPT-4-turbo into our study. This inclusion is made to validate the adaptability and robustness of our approach to DocRE across diverse model architectures.

We first leverage a 3-shot learning format for GPT mdoels to directly tackle DocRE, utilizing examples meticulously curated from the human-annotated DocRED dataset. These examples are carefully chosen to include a diverse range of entity pairs, capturing both relation-expressing and non-relation entity pairs, thereby mirroring the complexity and variability inherent in real-world documents. Then, we adopt the LMRC paradigm. In this setting, GPT models take over the role of LLaMA models in the RC stage, augmented by the addition of 3-shot learning format to provide essential examples and enrich contextual understanding. 

\begin{table}[ht]
\centering
\scalebox{0.63}{
\begin{tabular}{lcccc}
\toprule
\textbf{Methods} & Ign $F_1$ & $F_1$ & Intra $F_1$ & Inter $F_1$ \\
\midrule
GPT-3.5-turbo-16k (3-shot)       & 6.57  & 6.97 & 11.78 & 4.24 \\
GPT-4-turbo (3-shot)             & 9.85 & 10.71 & 14.52 & 7.43 \\
LMRC + GPT-3.5-turbo-16k (3-shot)        & 27.27 & 28.39 & 38.52 & 17.19 \\
LMRC + GPT-4-turbo (3-shot)              & 34.84 & 36.20 & 44.08 & 27.30 \\
\midrule
LMRC + LLaMA3.1-8B-Instruct (FT)      & 59.22 & 61.08 & 66.70 & 54.17 \\
\bottomrule
\end{tabular}
}
\caption{Performance of GPT models on sampled documents from the DocRED dev set. FT: fine-tuning.}
\label{tab:gpt4}
\end{table}

Considering budget constraints, our experimental analysis is conducted on a randomly selected subset of 100 documents from the DocRED development set. The preliminary results are shown in Table \ref{tab:gpt4}. 
Before implementing LMRC, the extraction performances of GPT models are quite low, while their overall F1 scores are significantly improved after integrating LMRC strategy. This reflects the inherent challenges posed by DocRE and also demonstrates the effectiveness of our proposed method on various models. In addition, even GPT-4-turbo failed to surpass the fine-tuned LLaMA3.1-8B-Instruct through 3-shot in-context learning, indicating the necessity of fine-tuning LLMs on DocRE tasks.

\section{Conclusion}
In this work, we investigate the underlying reasons for LLM's limited effectiveness in document-level relation extraction and introduce a new approach, Large Language Models with Relation Classifier (LMRC), for DocRE. Our method comprises two main stages: relation candidate proposal and relation classification. Through experiments conducted on DocRED and Re-DocRED, we demonstrate the effectiveness of our proposed LMRC approach. The results further reveal that LMRC holds strong competitive advantages over other existing LLM-based methods. Our innovative model establishes a new standard, indicating its potential as a viable framework for future DocRE research.

\label{sec:bibtex}
\bibliography{custom}

\appendix
\onecolumn
\section{Prompts}
\begin{tcolorbox}[title=Prompt for Document-level Relation Extraction, label=box:prompt_re]
Below is an instruction that describes a task, paired with an input that provides further context. Write a response that appropriately completes the request.\\
\\
\#\#\# Instruction: \\
Your task is to determine whether there are relations between the entity pairs based on the information in the text. If there exists relations, select relations for the entity pairs from the relation set; if there is no relation, return None. \\
The format of the input entity pair is ‘(head entity| -| tail entity)’. \\
Your output format is ‘(head entity| relation/None| tail entity)’. \\
\\
\#\#\# Relation set: \\
\{predefined relation set\} \\
\\
\#\#\# Text: \\
\{text\} \\
\\
\#\#\# \{number of entity pairs\} Entity pairs: \\
\{entity pairs\} \\
\\
\#\#\# Response:
\end{tcolorbox}

\vspace{5pt}

\begin{tcolorbox}[title=Prompt for Relation Classification, label=box:prompt_rc]
Below is an instruction that describes a task, paired with an input that provides further context. Write a response that appropriately completes the request.\\
\\
\#\#\# Instruction: \\
This is a relation classification task. we will provide entity pairs that require relation classification. Your task is to select relations for each entity pair from the given relation set based on the information in the text. There may be multiple relations between an entity pair. \\
The format of the input entity pair is ‘(head entity| -| tail entity)’. \\
Your output format is ‘(head entity| relation| tail entity)’. \\
\\
\#\#\# Relation set: \\
\{predefined relation set\} \\
\\
\#\#\# Text: \\
\{text\} \\
\\
\#\#\# \{number of entity pairs\} Entity pairs: \\
\{entity pairs\} \\
\\
\#\#\# Response: \\
\end{tcolorbox}

\section{Parameter settings}
\label{sec:para}
\begin{table}[ht]
\centering
{
\begin{tabular}{lcc}
\toprule
\textbf{Parameters} & \textbf{DocRED} &\textbf{Re-DocRED}\\
\midrule
batch size          &   4           & 4   \\
\# Epoch            &   20, \textbf{30}, 40    & \textbf{30}, 40\\
lr for encoder      &   \{5, \textbf{3}, 1\}e-5& \{\textbf{3}, 1\}e-5\\
lr for classifier   &   1e-4        & 1e-4\\
max gradient norm   &   1.0         & 1.0 \\
\bottomrule
\end{tabular}
}
\caption{Settings for the RCP stage.}
\label{tab:hyper_RCP}
\end{table}

\begin{table}[ht]
\centering
{
\begin{tabular}{lcccc}
\toprule
 \multirow{2}{*}{\textbf{Parameters}} & \multicolumn{2}{c}{\textbf{Pre}} & \multicolumn{2}{c}{\textbf{RC stage}} \\
 & \scriptsize\textbf{DocRED} & \scriptsize\textbf{Re-DocRED} & \scriptsize\textbf{DocRED} & \scriptsize\textbf{Re-DocRED} \\
\midrule
 batch size      & 4   & 4   & 4    & 4   \\
 \# Epoch        & 2   & 2   & 8    & 8   \\
 learning rate   & 1e-4& 1e-4& 1e-4 & 1e-4\\
 warmup steps    & 200 & 200 & 100  & 100 \\
 lora r          & 8   & 8   & 8    & 8   \\
 lora alpha      & 16  & 16  & 16   & 16  \\
\bottomrule
\end{tabular}
}
\caption{Settings for LoRA fine-tuning. (Pre stands for preliminary experiment)}
\label{tab:hyper_lora}
\end{table}

\end{document}